\documentclass[letterpaper, 10 pt, conference]{ieeeconf}  

\IEEEoverridecommandlockouts                              
\let\labelindent\relax

\usepackage[]{graphicx}    
\newcommand{\ignore}[1]{}  

\usepackage{amsfonts}
\usepackage{bm}
\usepackage{stackengine}
\usepackage{amssymb}
\usepackage{amsmath}
\usepackage{graphicx}
\usepackage{float}
\usepackage{hyperref}

\usepackage{wrapfig}
\usepackage{mwe}
\usepackage{paralist}
\usepackage[shortlabels]{enumitem}
\usepackage[utf8]{inputenc}
\usepackage{epstopdf}
\usepackage{array}
\usepackage{tabularx}
\usepackage{stfloats}
\usepackage{algorithm}
\usepackage{algpseudocode}
\usepackage{fancyhdr}
\usepackage{xcolor}
\usepackage{subcaption}
\usepackage{cite}

\newtheorem{proposition}{Proposition}

\title{\LARGE \bf
Data-Driven Risk-sensitive Model Predictive Control for Safe Navigation in Multi-Robot Systems
}

\author{Atharva Navsalkar$^{1}$ and Ashish R. Hota$^{2}$
\thanks{This work is supported by AI \& Robotics Technology Park (ARTPARK), Indian Institute of Science (IISc), Bangalore, India through Student Innovation Grant Programme and by IIT Kharagpur under the ISIRD scheme.}
\thanks{$^{1}$Atharva Navsalkar is with the Department of Mechanical Engineering, Indian Istitute of Technology (IIT) Kharagpur, India. {\tt\small Email: anavsalkar@iitkgp.ac.in}.}%
\thanks{$^{2}$Ashish R. Hota is with the Department of Electrical Engineering, Indian Institute of Technology (IIT) Kharagpur, India. {\tt\small Email:  ahota@ee.iitkgp.ac.in}.}%
}

\begin{document}

\maketitle
\thispagestyle{empty}
\pagestyle{empty}

\begin{abstract}
Safe navigation is a fundamental challenge in multi-robot systems due to the uncertainty surrounding the future trajectory of the robots that act as obstacles for each other. In this work, we propose a principled data-driven approach where each robot repeatedly solves a finite horizon optimization problem subject to collision avoidance constraints with latter being formulated as distributionally robust conditional value-at-risk (CVaR) of the distance between the agent and a polyhedral obstacle geometry. Specifically, the CVaR constraints are required to hold for all distributions that are close to the empirical distribution constructed from observed samples of prediction error collected during execution. The generality of the approach allows us to robustify against prediction errors that arise under commonly imposed assumptions in both distributed and decentralized settings. We derive tractable finite-dimensional approximations of this class of constraints by leveraging convex and minmax duality results for Wasserstein distributionally robust optimization problems. The effectiveness of the proposed approach is illustrated in a multi-drone navigation setting implemented in Gazebo platform.
\end{abstract}

\section{Introduction}
\label{sec:introduction}
Multi-robot systems, including drones, ground robots, and autonomous vehicles, have seen tremendous growth due to applications ranging from military \cite{gans2021cooperative}, search and rescue missions \cite{queralta2020collaborative}, advanced mobility \cite{malik2021collaborative}, cave explorations \cite{rouvcek2019darpa}, indoor motion \cite{jiang2019multi}, warehouses \cite{pinkam2016robot} and entertainment purposes \cite{o2021there}. In such settings, each individual robot moves in a highly uncertain environment and is required to safely avoid obstacles and other members of the group. Therefore, motion planning in uncertain environments, guaranteeing safety in terms of collision avoidance and distributed computation are some of the key challenges in autonomous multi-robot systems; see \cite{huang2019collision} for a recent review. In this context, optimization based motion planning strategies, including Model Predictive Control (MPC), are increasingly being considered for safe navigation of robotic systems \cite{luis2020online,kamel2017robust,zhang2020optimization,batkovic2020robust,soloperto2019collision,katriniok2018distributed,dixit2022risk,cheng2017decentralized,hakobyan2019risk}.

In particular, MPC is a powerful framework that repeatedly solves a finite horizon numerical optimization problem to compute control commands that optimize suitable performance metrics while satisfying (collision avoidance) constraints. Collision avoidance constraints are often modeled as distance of the controlled object from the obstacle being larger than a safe limit. In the single-agent setting, these constraints are deterministic for static obstacles \cite{zhang2020optimization}, while the presence of uncertainty and dynamic obstacles results in these constraints being stochastic in nature. The authors in \cite{soloperto2019collision} consider robust constraint satisfaction leading to highly conservative trajectories. More recent works such as \cite{castillo2020real,zhu2019chance,batkovic2020robust,de2021scenario} leverage probability distribution of the uncertainty or past samples to guarantee that collision avoidance constraints are satisfied with high probability. Authors in \cite{gao2021risk,soloperto2019collision} focus primarily on autonomous driving scenarios. 

The problem is more challenging in multi-robot systems where other mobile robots act as dynamic obstacles for the controlled agent. There are two paradigms in this context.
\begin{itemize}
    \item {\it Distributed setting} where it is assumed that each agent solves its own MPC problem, communicates the computed trajectory with neighboring agents, and formulates collision avoidance constraints in terms of the future position of neighboring agents received from them (see  \cite{firoozi2020distributed,luis2020online} for deterministic, \cite{dai2022distributed} for robust and \cite{katriniok2018distributed} for chance constrained formulations).
    \item {\it Decentralized setting} where there is no communication, each agent predicts the future position of neighboring agents, often by assuming that other agents will continue to move at their present velocity, and avoids collision with the predicted position of other agents (see \cite{arul2020dcad,berg2011reciprocal,cheng2017decentralized} for deterministic, \cite{park2020online,kamel2017robust} for robust, and \cite{gopalakrishnan2017prvo,arul2021swarmcco,zhang2021receding} for chance constrained approaches).
\end{itemize}
Most of the above works consider both the controlled agent and obstacles as point mass entities and do not consider the detailed geometry of the obstacles. The assumption regarding future position of other agents may not hold during execution as agents continuously update their strategies (e.g., velocity does not remain constant over the horizon). Furthermore, robust optimization approaches provide highly conservative solutions while chance constraints do not provide any guarantees on the magnitude of constraint violation in the (less likely) event that there is a collision. 

In order to provide such guarantees while being less conservative, it is more appropriate to formulate the uncertain safety (collision avoidance) constraints in terms of coherent risk measures such as {\it conditional value-at-risk} (CVaR) \cite{wang2022risk}. The authors in \cite{hakobyan2019risk} propose CVaR based motion planning when obstacle motion is affected by Gaussian randomness. In a follow up work \cite{hakobyan2021wasserstein}, a distributionally robust formulation is proposed which guarantees that the CVaR of the collision avoidance constraint is satisfied for an entire family of distributions constructed from past observations. A similar setting is also studied in \cite{dixit2022risk}. While these works model obstacles as polyhedrons, collision avoidance constraints considered there do not distinguish between solutions when collision does not take place, leading to less safety margins and possibly risky maneuvers in the computed trajectories. In addition, the above works are limited to the single-agent setting. 

Table \ref{tab_new} summarises the previous works in this domain and highlights the key research gap in the literature that there are few approaches that have investigated risk sensitive (CVaR-based) collision avoidance constraints in multi-robot systems in a principled manner that takes into account obstacle geometry, error in prediction of future position of other agents, and unknown distribution of uncertain parameters.

\begin{table}[tbp]
\centering
\caption{Classification of previous works}
\label{tab_new}
\begin{tabular}{|p{0.17\linewidth}|p{0.17\linewidth}|p{0.1\linewidth}|p{0.16\linewidth}|p{0.14\linewidth}|}
\hline 
Computation & Deterministic & Robust & Chance constraints & CVaR\\\hline\hline
Single-agent & \cite{zhang2020optimization} & \cite{soloperto2019collision} & \cite{castillo2020real,zhu2019chance} \cite{batkovic2020robust,de2021scenario} & \cite{gao2021risk,hakobyan2021wasserstein,dixit2022risk} \\\hline
Decentralised & \cite{arul2020dcad,berg2011reciprocal,cheng2017decentralized} & \cite{park2020online,kamel2017robust} & \cite{gopalakrishnan2017prvo,arul2021swarmcco,zhang2021receding} & {\bf Our work} \\\hline
Distributed & \cite{firoozi2020distributed,luis2020online} & \cite{dai2022distributed} & \cite{katriniok2018distributed} & {\bf Our work} \\\hline
\end{tabular}
\end{table}

\textbf{Our contributions:} We consider a multi-robot system where each robot solves a MPC problem subject to constraints on collision avoidance. At each time, an agent collects samples of the prediction error between the current position of other robots and the position of the other robots predicted in the past in the decentralized setting or shared by the other robots in the past in the distributed setting. The collision avoidance constraints are then formulated as distributionally robust CVaR constraints on the distance between the controlled object and a polyhedral obstacle parameterized by the predicted position and the uncertain prediction error. In other words, we require the CVaR of the distance to be bounded for a family of distributions of the uncertain prediction error close (in the sense of Wasserstein metric) to the past samples of prediction errors. While this class of constraints are infinite-dimensional, we derive tractable finite-dimensional approximations by leveraging convex and minmax duality results for distributionally robust optimization problems. Finally we demonstrate the efficacy of the proposed approach in a multi-drone navigation setting implemented in Gazebo platform with multi-agent MPC being executed in parallel processors with realistic inter-agent communication protocols in place.

\section{Problem Definition and Solution Approach}
\label{sec:system_description}
We consider a multi-agent system comprising of $N$ individual mobile robots or agents. The goal of each agent is to reach a (agent-specific) final position or track a desired trajectory. For both objectives, the cost function for an agent $i$ at time $k+l$ computed at time $k$ is given by
\begin{align}
    J_i(k+l|k) & = \mathbf{x}_{e,i}(k+l|k)^{\intercal}Q\;\mathbf{x}_{e,i}(k+l|k) \nonumber
    \\ & \qquad \quad + u_i(k+l|k)^{\intercal}Ru_i(k+l|k),
    \label{eq:stagecost}
\end{align}
where $\mathbf{x}_{e,i}(k+l|k) = \mathbf{x}_{\text{ref},i}(k+l) -  \mathbf{x}_i(k+l|k)$ is the difference between the desired state at time $k+l$ and the state at time $k+l$ predicted at $k$. The second term penalises the control effort with $u_i(k+l|k)$ being the control input for time $k+l$ computed at time $k$, and the matrices $Q$ and $R$ are assumed to be positive definite. The finite horizon optimal control problem for agent $i$ is given by 
\begin{equation}\label{eq:OCP}
    \begin{aligned}
    \min_{\mathbf{x}_i,\mathbf{u}_i} \quad &  \sum_{l=1}^{T} J_i(k+l|k) \\
    \text{s.t.} \quad & x_i(k+l|k) = f_i(x_i(k+l-1|k), u_i(k+l-1|k)), \\
    & x_i(k+l|k) \in \mathcal{X}_i, \; u_i(k+l-1|k) \in \mathcal{U}_i,\\
    & \mathcal{C}(z_i(k+l|k),z_j(k+l|k)) \leq 0, \quad \forall i \neq j, \\
    & \text{for all } l \in [T], j \in [N],
    \end{aligned}
\end{equation}
where $f_i$ captures the discrete-time dynamics, $\mathcal{X}_i$ and $\mathcal{U}_i$ denote the deterministic constraints on states and control inputs for agent $i$, and $\mathcal{C}(z_i,z_j) \leq 0$ denote the collision avoidance constraints between two agents $i$ and $j$ with positions $z_i$ and $z_j$, respectively. The position is assumed to be part of the state vector. The above problem is solved for states $\mathbf{x}_i$, control inputs $\mathbf{u}_i$ for agent $i$ at each time step. For ease of notation, we define $[N] := \{1,...,N\}$. Thus, in order to ensure safe navigation of agent $i$, we need to compute optimal control inputs such that agent $i$ does not collide with any other agent $j \neq i$ over the prediction horizon. 

\subsection{Obstacle Occupancy Modeling}

From the perspective of agent $i$, other agents act as obstacles which occupy some space that is forbidden for it. The occupancy set is modelled as convex polyhedral sets composed as union of multiple half-spaces. In particular, the space occupied by obstacle $m$ is represented as 
\begin{equation}
    \mathcal{O}^m = \{p \in \mathbb{R}^3: A^m p\leq b^m\}, \;\; m \in [M],
    \label{eq:obstdef}
\end{equation}
where $M$ denotes the total number of obstacles, $A^m \in \mathbb{R}^{n_m \times 3}$ and $b^m \in \mathbb{R}^{n_m}$ are constant matrices and vectors that represent the position and orientation of the obstacle, and $n_m$ is the number of half spaces required to model obstacle $m$. We assume that any non-convex obstacle can be conservatively approximated to an enclosed polyhedron.

As the agents are considered to be dynamic obstacles, the polyhedral representation of each obstacle is also a function of time. As a result, agent $i$ needs to know the predicted occupancy sets of all other agents $T$ steps into the future. This information is not readily available. As discussed earlier, there are two main paradigms in the literature based on how this information is accessed. In the {\it distributed approach}, agents exchange their optimal solution (future trajectory) with others. Thus, at each time step $k$, agent $i$ receives the presently computed MPC solution which includes future position and orientation information of other agents. However, other agents may not follow the current optimal trajectory and as a result, this approach is not robust to this future deviation by other agents. In the {\it decentralized approach}, there is no inter-agent communication, and most prior works assume that agent $i$ predicts the future position of other agents assuming that other agents will continue to move with their present velocity. Thus, this assumption is rather naive and often leads to incorrect predictions and collisions. 

In this work, we propose a principled approach to robustify against such prediction inaccuracies in both distributed and decentralized schemes. We start by assuming that at time $k$, the controlled agent $i$ has access to anticipated future position of other agents over the prediction horizon, i.e., it is aware of $z_j(k+l|k)$ for $l \in [T], j \in [N]$. In the distributed case, this information is shared by other agents and corresponds to the solution of their MPC problem at the previous time step. In the decentralized case, this information is predicted by the agent under the constant velocity assumption. With regard to orientation, we assume that the present occupancy set of an obstacle agent $j$, denoted by $\mathcal{O}^j_{k|k}$ and characterized by $A^j(k)$ and $b^j(k)$, is known to agent $i$, and the future orientation of agent $j$ remains unchanged from its present orientation. Under the above assumption, the uncertain obstacle space of agent $j$ predicted at time $k$ for time step $k+l$ is formally stated as
\begin{equation}
    \mathcal{O}^j_{k+l|k}=\mathcal{O}^j_{k|k} + z_j(k+l|k)+w_j^{(l)}, 
    \label{eq:obst@l}
\end{equation}
where $w_j^{(l)} \in \mathbb{R}^3$ is the difference between the true position and the anticipated position of this agent $l$ steps into the future. A similar set up with linearly perturbed uncertainty sets was considered in \cite{hakobyan2019risk} in the single-agent case. 

While the probability distribution of $w_j^{(l)}$ is unknown, the controlled agent has access to samples of $w_j^{(l)}$ from past trajectory as follows: at each time step $k$, when position of an agent $j$ is observed, we compare it with the predictions of the position of agent $j$ obtained in previous $T$ time steps and compute the difference as samples of the the uncertain parameter $w_j^{(l)}, l \in [T]$. In other words, $z_j(k|k) - z_j(k|k-q)$ is treated as a sample of $w_j^{(q)}$ which is the $q$-step prediction error. Thus, at time $k$, we collect a set of samples of $w_j^{(l)}, \quad l \in [T]$ as described above. We now describe the formulation of data-driven distributionally robust collision avoidance constraints. 


\subsection{Collision avoidance constraint formulation}

Given the uncertain occupancy set defined in \eqref{eq:obst@l}, the collision avoidance constraint is now stated as
\begin{align}
    & F\big(z_i(k+l|k), z_j(k+l|k), w_j^{(l)} \big) := \nonumber
    \\ & \qquad \quad d_{\text{min}} - \text{dist}\big( z_i(k+l|k),\;  \mathcal{O}^j_{k+l|k}\big) \leq 0, \label{eq:fdef}
\end{align}
where the {\it dist} function is the distance between the agent position $z_i(k+l|k)$ and obstacle space $\mathcal{O}^j_{k+l|k}$. The above constraint is required to hold for all neighbors $j \in [N]$ and time $l \in [T]$ in the MPC problem of agent $i$ at time $k$. 

\subsubsection{Reformulation in the deterministic setting}

Before introducing the distributionally robust risk sensitive version of the above constraints, we first present the reformulation of the above in the deterministic regime. Consider the occupancy set $\mathcal{O}^m$ defined in \eqref{eq:obstdef}. The distance between an agent at position $z_i$ and $\mathcal{O}^m$ is given by
\begin{equation}
\begin{aligned}
    & \text{dist}(z_i,\mathcal{O}^m) := \min_{r \in \mathcal{O}^m} ||z_i-r|| \; \\
    & \qquad = \min_d(||d||: A^m(z_i+d) \leq b^m).
    \label{eq:DistanceDefinition}
\end{aligned}
\end{equation}
It is evident that the constraint \eqref{eq:fdef} with the above definition of distance is non-trivial to impose on the optimization problem since the distance function itself involves solving an optimization problem. The following result from \cite{zhang2020optimization} proposes an equivalent tractable form for these constraints by leveraging convex duality. 

\begin{proposition}[\cite{zhang2020optimization}]\label{prop:obca_dual}
For an obstacle set $\mathcal{O} = \{p \in \mathbb{R}^3: A^m p \leq b^m\}$, we have 
\begin{align}
    & \text{dist}(z_i,\mathcal{O}^m) \geq 0 \Longleftrightarrow \nonumber
    \\ & \qquad \exists \lambda \geq 0: (A^m z_i-b^m)^{\intercal} \lambda \geq 0,\; ||(A^m)^{\intercal}\lambda||_2 \leq 1.
    \label{eq:reform}
\end{align}
\end{proposition}

\vspace{2mm}

Thus, if there exists $\lambda$ satisfying the above constraints, then the collision constraint is satisfied (the distance between the controlled agent and the obstacle is non negative). As a result, these conditions can be encoded as constraints with $\lambda$ being an additional decision variable in the MPC formulation. We now introduce the distributionally robust framework to appropriately handle the uncertain constraint \eqref{eq:fdef}.
\subsection{Data-Driven Distributionally Robust Constraint Formulation}

Note that the constraint function $F(z_i,z_j,w_j)$ defined in \eqref{eq:fdef} is uncertain with the distribution of $w_j$ not being known. However, a collection of $N_s$ samples of $w_j$ is available with MPC controller of agent $i$ denoted by $\{\hat{w}_{j,n}\}_{n \in [N_s]}$. We leverage these available samples to define data-driven distributionally-robust conditional value-at-risk (CVaR) constraints on the function $F$ as
\begin{equation}
    \sup_{\mathbb{P}\in\mathcal{M}^\theta_{N_s}} \text{CVaR}^{\mathbb{P}}_{1-\alpha} \left[ F\left( z_i, z_j, w_j \right) \right] \leq 0, \label{eq:cvar_cons}
\end{equation}
where 
\begin{itemize}
    \item the CVaR of a random loss $X$ with distribution $\mathbb{P}$, is equal to the conditional expectation of the loss within the $\alpha$ worst case quantile of the loss distribution, i.e.,  
    \begin{equation}
        \text{CVaR}_{1-\alpha}^{\mathbb{P}}(X) := \inf_{z \in \mathbb{R}}\left[\alpha^{-1}\mathbb{E}[(X+z)^{+}]-z\right],
        \label{eq:cvar}
    \end{equation}
    where $(x)^{+} = \max\{x,0\}$. Consequently, CVaR constraint aims to constrain the value at the tail distribution. 
    \item the set $\mathcal{M}^\theta_{N_s}$ is a family of probability distributions that are within a Wasserstein distance $\theta$ from the empirical distribution induced by the $N_s$ samples $\{\hat{w}_{j,n}\}_{n \in [N_s]}$; the formal definition of the ambiguity set is omitted in the interest of space and can be found in \cite{hota2019data,hakobyan2021wasserstein}. 
\end{itemize}

Following the definition of CVaR, the constraint \eqref{eq:cvar_cons} assumes the form: 
\begin{equation}\label{eq:cvar_cons2}
    \sup_{\mathbb{P}\in\mathcal{M}^\theta_{N_s}}  \inf_{t\in\mathbb{R}} \mathbb{E}^{\mathbb{P}}\left[\left( F\left( z_i, z_j, w_j \right) + t\right)^+ - t\alpha\right] \leq 0.
\end{equation}
The above constraint is infinite-dimensional due to the supremum being over a family of probability distributions. In the remainder of this subsection, we approximate and reformulate the above constraint into a finite-dimensional constraint which can be solved via off-the-shelf solvers.

First we observe that since $\left(\sup\;\inf\; [\;\;]\right) \leq \left(\inf\;\sup\;[\;\;]\right)$, the constraint 
\begin{equation}
     \inf_{t\in\mathbb{R}}
     \sup_{\mathbb{P}\in\mathcal{M}^\theta_{N_s}}\mathbb{E}^{\mathbb{P}}\left[\left( F\left( z_i, z_j, w_j \right) + t\right)^+ - t\alpha\right] \leq 0.
\end{equation}
is sufficient for \eqref{eq:cvar_cons2} to hold true. Now, the inner supremum problem in the above equation can be reformulated as shown in \cite{hota2019data} to an infimum problem, which then combined with the the infimum over $t$ yields the following set of constraints that are sufficient for \eqref{eq:cvar_cons2} to hold true:
\begin{subequations}
\begin{align}
    &\lambda_{\theta}\theta - t\alpha + \frac{1}{N_s}\sum_{n=1}^{N_s}s_n \leq 0, \\
    &s_n \!\geq \!\sup_{w_j \in \Omega_j} [ F\left( z_i, z_j, w_j \right)+\!t-\!\lambda_{\theta}\! ||w_j-\!\hat{w}_{j,n}||_{2} ], n \in [N_s], \label{eq:partb}\\
    & s_n \geq 0,\;\;\;t\in\mathbb{R},\;\;\;\lambda_{\theta}\geq0, \nonumber
\end{align}
\end{subequations}
where $\theta$ is the radius of the ambiguity set. We now focus on reformulating the semi-infinite constraint \eqref{eq:partb} which involves an optimization problem over the support of $w_j$ denoted by $\Omega_j$ in the following two major steps. 

\noindent {\bf Step 1: Reformulation of \eqref{eq:partb}.}

From the definition of the constraint function $F$ in \eqref{eq:fdef}, we express \eqref{eq:partb} for sample $n$ as
\begin{align}
    s_n & \geq \sup_{w_j\in \Omega_j} \Bigr[ d_{\text{min}}-\text{dist}\left( z_i, \mathcal{O}^j \right) +t-\lambda_{\theta}||w_j-\!\hat{w}_{j,n}||_{2} \Bigr] \nonumber\\
    & = d_{\text{min}} + t - \inf_{w_j \in \Omega_j} \Bigr[ \text{dist}\left( z_i, \mathcal{O}^j \right) +\lambda_{\theta}||w_j-\!\hat{w}_{j,n}||_{2} \Bigr] \nonumber.
\end{align}

Based on the representation \eqref{eq:obstdef}, when $\mathcal{O}^j_{k|k} = \{ p \in \mathbb{R}^3|Ap \leq b\}$, then the distance function in the above equation is given by 
\begin{align}
    \min & \;\;\;||t|| \nonumber\\
    \text{s.t.} & \;\;\; A\left(z_i+t-z_j-w_j\right) \leq b,
\end{align}
and following the strong duality result in Proposition \ref{prop:obca_dual}, it can be stated equivalently as 
\begin{align}
    \max_{\mathbf{\lambda}\geq 0} & \; \left[A(z_i-z_j-w_j)-b\right]^{\intercal}\mathbf{\lambda}\nonumber \\
    \text{s.t.} & \; ||A^\intercal \lambda||_2 \leq 1.  \label{eq:eqform}
\end{align}
Substituting the above in the inequality involving $s_n$ yields
\begin{align}\label{eq:int_minmax}
    s_n \geq & d_{\text{min}} + t -  
    \inf_{w_j \in \Omega_j} \Bigr[ 
     \max_{\lambda \geq 0,\;||A^{\intercal}\lambda||_2 \leq 1 }\bigl\{ [A(z_i-z_j-w_j) \nonumber
     \\ &\quad
     -b]^{\intercal}\mathbf{\lambda}\bigr\}
     +\lambda_{\theta}||w_j - \hat{w}_{j,n}||_2 \Bigr].
\end{align}

Once again, note that since $\left(\sup\;\inf\; [\;\;]\right) \leq \left(\inf\;\sup\;[\;\;]\right)$, the inequality 
\begin{align}
    s_n \geq & d_{\text{min}} + t - \max_{\lambda \geq 0,\;||A^{\intercal}\lambda||_2 \leq 1 } 
     \Bigr[ \inf_{w_j \in \Omega_j}
     \bigl\{ \bigl[A(z_i-z_j-w_j) \nonumber
     \\ &\quad
     -b\bigr]^{\intercal}\mathbf{\lambda}
     +\lambda_{\theta}||w_j - \hat{w}_{j,n}||_2 \bigr\} \Bigr],
\end{align}
is sufficient for \eqref{eq:int_minmax} to hold.

Rearranging the equations, we obtain
\begin{align}
  s_n \geq &d_{\text{min}} + t - 
  \max_{\lambda \geq 0,\;||A^{\intercal}\lambda||_2 \leq 1 } \Bigr([A(z_i-z_j)-b]^{\intercal}\mathbf{\lambda}  \nonumber\\ 
  &\quad\quad
  +\inf_{w_j \in \Omega_j} \bigr[ \lambda_{\theta}||w_j - \hat{w}_{j,n}||_2-w_j^\intercal(A^{\intercal}\lambda) \bigr]\Bigr) \label{eq:rearranging}.
\end{align}

\noindent {\bf Step 2: Reformulation of the infimum with respect to $w_j$.}

The infimum term with respect to $w_j$ can be written as
\begin{align}
    & \inf_{w_j \in \Omega_j} \bigr[- \left( \lambda^{\intercal}Aw_j - \lambda_{\theta}||w_j - \hat{w}_{j,n}||_2 \right) \bigr],\\
    \iff & -\sup_{w_j \in \Omega_j} \bigr[ \lambda^{\intercal}Aw_j - \lambda_{\theta}||w_j - \hat{w}_{j,n}||_2  \bigr]. \label{eq:inner3}
\end{align}

When the support of $w_j$ is a polyhendron, i.e., $\Omega_j = \{w \in \mathbb{R}^3|C_j w \leq h_j\}$, the authors in \cite{hota2019data} showed that the supremum term above is equivalent to
\begin{align}
    \min_{\eta_{j,n} \geq 0} & \qquad \; (A^\intercal \lambda - C_j^\intercal \eta_{j,n})^\intercal \hat{w}_{j,n} + \eta^\intercal_{j,n} h_j  \nonumber \\
    \text{s.t.} & \qquad \; ||A^\intercal \lambda - C_j^\intercal \eta_{j,n}||_2 \leq \lambda_\theta.  \label{eq:eqform}
\end{align}
As a result, \eqref{eq:inner3} can be stated equivalently as
\begin{align}
    \max_{\eta_{j,n} \geq 0} & \qquad \; -\bigl[(A^\intercal \lambda - C_j^\intercal \eta_{j,n})^\intercal \hat{w}_{j,n} + \eta^\intercal_{j,n} h_j\bigr]  \nonumber \\
    \text{s.t.} & \qquad \; ||A^\intercal \lambda - C_j^\intercal \eta_{j,n}||_2 \leq \lambda_\theta.  \label{eq:inner4}
\end{align}

Consequently, \eqref{eq:rearranging} can be stated equivalently as
\begin{align}
  s_n \geq & d_{\text{min}} + t - 
  \max_{\lambda \geq 0,\;||A^{\intercal}\lambda||_2 \leq 1 } \Bigr([A(z_i-z_j)-b]^{\intercal}\mathbf{\lambda}  \nonumber\\ 
  & \!\!+\!\!\!\!\! \max_{\substack{\eta_{j,n} \geq 0, \\ ||A^\intercal \lambda - C_j^\intercal \eta_{j,n}||_2 \leq \lambda_\theta}} \!\!\! -[(A^\intercal \lambda - C_j^\intercal \eta_{j,n})^\intercal \hat{w}_{j,n} + \eta^\intercal_{j,n} h_j] \Bigr) \label{eq:rearranging2}.
\end{align}

Since the maximum terms on the R.H.S are preceded by a negative sign, the following constraints are sufficient to guarantee that the constraint in \eqref{eq:rearranging2} holds:
\begin{subequations}
\begin{align}
  & s_n \geq d_{\text{min}} + t - \biggr([A(z_i-z_j)-b]^{\intercal}\mathbf{\lambda} \nonumber
    \\ & \qquad -[(A^\intercal \lambda - C_j^\intercal \eta_{j,n})^\intercal \hat{w}_{j,n} + \eta^\intercal_{j,n} h_j] \biggr), 
  \\ & \lambda \geq 0, \quad ||A^{\intercal}\lambda||_2 \leq 1,
  \\ & \eta_{j,n} \geq 0, \quad ||A^\intercal \lambda - C_j^\intercal \eta_{j,n}||_2 \leq \lambda_\theta.
\end{align}
\end{subequations}

When the support $\Omega_j$ is not known and is assumed to be $\mathbb{R}^3$, we do not need the multipliers $\eta_{j,n}$, and consequently, the following set of constraints 
\begin{subequations}
\begin{align}
  & s_n \geq d_{\text{min}} + t - \biggr([A(z_i-z_j)-b]^{\intercal}\mathbf{\lambda} -[(A^\intercal \lambda)^\intercal \hat{w}_{j,n}] \biggr), 
  \\ & \lambda \geq 0, \quad ||A^{\intercal}\lambda||_2 \leq 1, \quad ||A^\intercal \lambda||_2 \leq \lambda_\theta.
\end{align}
\end{subequations}
are sufficient to guarantee that \eqref{eq:rearranging2} holds. 

To summarize, the original distributionally CVaR collision avoidance constraint \eqref{eq:cvar_cons} can be approximated as
\begin{subequations}\label{eq:drcvar_final}
\begin{align}
    &\lambda_{\theta}\theta - t\alpha + \frac{1}{N_s}\sum_{n=1}^{N_s}s_n \leq 0, \\
    & d_{\text{min}} + t - s_n \leq [A(z_i-z_j)-b]^{\intercal}\mathbf{\lambda} -[ \lambda^{\intercal}A\hat{\omega}_{j,n}]\\
    & \lambda \geq 0,\;||A^{\intercal}\lambda||_2 \leq \min(1,\lambda_\theta), \;\;\; 
    \\ & \lambda_{\theta}\geq0,\qquad t\in\mathbb{R}, \qquad s_n \geq 0 \quad \forall n \in [N_s]. 
\end{align}
\end{subequations}
Thus, the MPC problem for agent $i$ has the above set of constraints for each neighbor $j$ with $z_i,z_j, A, b$ being replaced by $z_i(k+l|k),z_j(k+l|k), A^j(k), b^j(k)$ for all time steps over the horizon $l \in [T]$.

\section{Simulation Results}
\label{sec:results}
In order to illustrate the effectiveness of the proposed approach, we consider a multi-drone system. Following \cite{sabatino2015quadrotor}, the nonlinear dynamics of the drone can be represented in the standard $\mathbf{\dot{x}} = f(\mathbf{x}, \mathbf{u})$ form as:
 \begin{equation}
     \begin{bmatrix} \dot{x} \\ \dot{y} \\ \dot{z} \\ \ddot{x} \\ \ddot{y} \\ \ddot{z} \\ \dot{\phi} \\ \dot{\theta} \\ \dot{\psi} \\ \dot{p} \\ \dot{q} \\ \dot{r} \end{bmatrix} = \begin{bmatrix} \dot{x} \\ \dot{y} \\ \dot{z} \\ \frac{1}{m}(c\psi s\theta + c\theta s\phi s\psi) u_1 \\ \frac{1}{m}(s\psi s\theta - c\theta s\phi c\psi) u_1 \\ \frac{1}{m}(c\phi c\theta ) u_1 - g \\ p(c\theta) + r(s\theta) \\ p(s\theta s\phi / c\phi) + q - r(c\theta s\phi / c\phi) \\ -p(s\theta/c\phi) + r(c\theta / c\phi) \\     \frac{1}{I_{xx}}(u_2 - (I_{zz} - I_{yy})qr) \\ \frac{1}{I_{yy}}(u_3 - (I_{xx} - I_{zz})pr) \\ \frac{1}{I_{zz}}(u_4 - (I_{yy} - I_{xx})pq) \end{bmatrix}, \label{eq:nonlin}
\end{equation}
where the state contains position coordinates ($x,y,z$), velocities ($\dot{x},\dot{y},\dot{z}$), Euler angles ($\phi,\theta,\psi$) for orientation, and angular velocities ($p,q,r$), and inertia $I$ has only the diagonal components $I_{xx}$, $I_{yy}$, $I_{zz}$. We denote $c\phi := \cos(\phi)$ and $s\phi := \sin(\phi)$ and so on, for better readability. The control input is $\mathbf{u} = [u_1,u_2,u_3,u_4]$, where $u_1$ is the thrust, and $u_2$, $u_3$, $u_4$ are the angular moments (these are transformed by a low-level controller into actuation signals). 

First, we use numerical simulator with nonlinear dynamics discretized with sampling time $1$ ms. The sampling time for MPC is chosen to be $0.1$ s. In this setup, all agents solve the MPC problem synchronously and the numerical simulator gives the next state of the drone. All these independent processes run on a workstation with AMD Ryzen $5800$H chipset and $16$GB RAM. In particular, we use Python \texttt{multiprocessing} module to launch parallel nodes for each agent interacting with the common simulation. The odometry data as well as predictions are communicated among each other using Robot Operating System (ROS). The MPC problem is solved using a nonlinear programming solver IPOPT \cite{wachter2006implementation}. Our implementation uses the MA27 as the linear solver for IPOPT. We use the ``do-mpc" interface \cite{lucia2017rapid} for the solver, that also uses CasADi package \cite{Andersson2019}.

Subsequently, for more realistic dynamics, we use Gazebo physics simulations for the AsTec Firefly hexacopter model using RotorS for low-level control \cite{furrer2016rotors}. Gazebo runs on the PC interacting with each agents for actuator commands and returning the odometry data.

\subsection{Distributed Setting}

In this subsection, we demonstrate our proposed formulation on numerical simulations in a distributed setting. To simplify the analysis, we start with a case with two agents, trying to cross each other on a straight path. Each agent considers itself to be a point mass, whereas the surrounding obstacles are assumed to be polyhedral. Each obstacle/agent is assumed to be a rectangular prism with a square cross-section of $2$ m. The length of prism is assumed very high to visualise planar collision avoidance. Unless stated otherwise, all subsequent results are obtained with a sample size of $N_s = 10$ and the prediction horizon of $T = 20$ steps.

\begin{figure*}[ht]
\centering
\begin{minipage}[b]{0.3\linewidth}
\centering
\includegraphics[width=\textwidth]{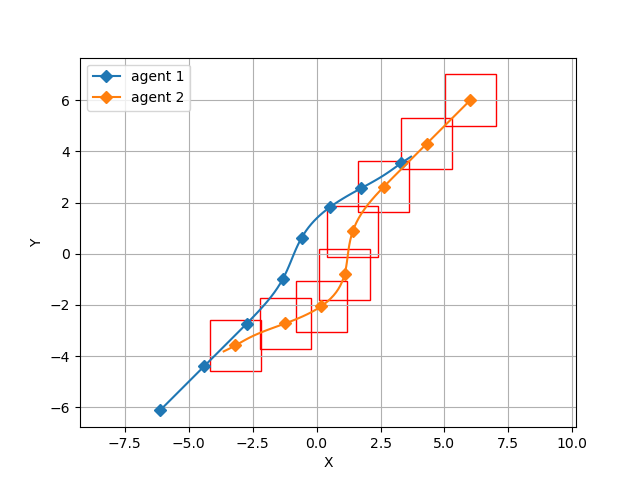}
\caption{Trajectory as seen by agent 1 for $\alpha=0.1$ and $\theta=0.001$.} 
\label{fig:rep_traj}
\end{minipage}
\hspace{5mm}
\begin{minipage}[b]{0.3\linewidth}
\centering
\includegraphics[width=\textwidth]{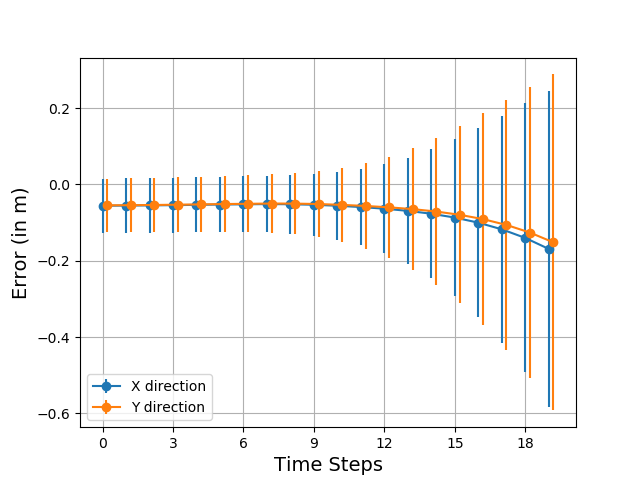}
\caption{Prediction errors across time horizon.}
\label{fig:alpha_errT}
\end{minipage}
\hspace{5mm}
\begin{minipage}[b]{0.3\linewidth}
\centering
\includegraphics[width=\textwidth]{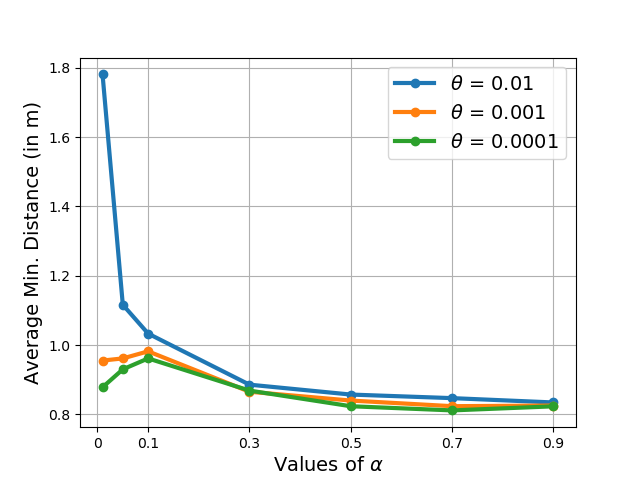}
\caption{Average minimum distance with increasing $\alpha$ values.}
\label{fig:alpha_dist}
\end{minipage}
\end{figure*}

Figure \ref{fig:rep_traj} shows the trajectory of both agents when the risk tolerance parameter (of the CVaR function) $\alpha = 0.1$ and the Wasserstein radius $\theta = 0.001$. The mean and standard deviation in errors in the predictions, i.e, the actual position of an obstacle agent and the optimal MPC solution of that agent $l$ steps before, is illustrated in Figure \ref{fig:alpha_errT}. The figure shows that deviations increases across the time horizon and does not necessarily have zero mean. Therefore, it is necessary to robustify the trajectories against these errors. 

Figure \ref{fig:alpha_dist} depicts the average value of the minimum distance between the agents over $50$ runs. Higher values of $\theta$ result in larger ambiguity sets around the collected samples which leads to more robust trajectories; this is observed from the figure which shows that average minimum distance is larger when $\theta$ is larger. As $\alpha$ increases, agents are more tolerant towards risk of collision, and as a result opt for risky trajectories which lead to reduced value of average minimum distance. The behaviour is more sensitive to $\theta$ for smaller values of $\alpha$. 

\begin{figure*}[ht]
\centering
\begin{minipage}[b]{0.3\linewidth}
\centering
\includegraphics[width=1\linewidth]{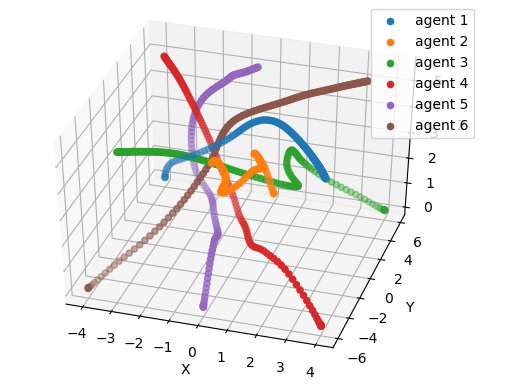}
\caption{3D visualisation of the Gazebo mission for $\alpha=0.1$ and $\theta=0.001$.}
\label{fig:hexa_3d}
\end{minipage}
\hspace{5mm}
\begin{minipage}[b]{0.3\linewidth}
\centering
\includegraphics[width=\textwidth]{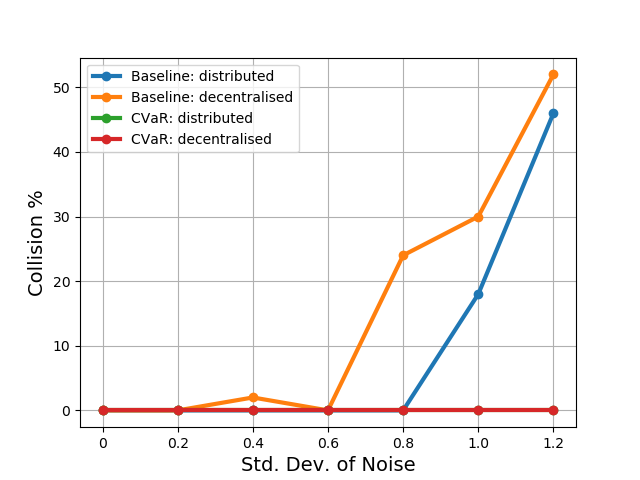}
\caption{Percentage of collisions with increasing noise values ( $\alpha=0.05$ and $\theta=0.001$).}
\label{fig:noise_percent}
\end{minipage}
\hspace{5mm}
\begin{minipage}[b]{0.3\linewidth}
\centering
\includegraphics[width=\textwidth]{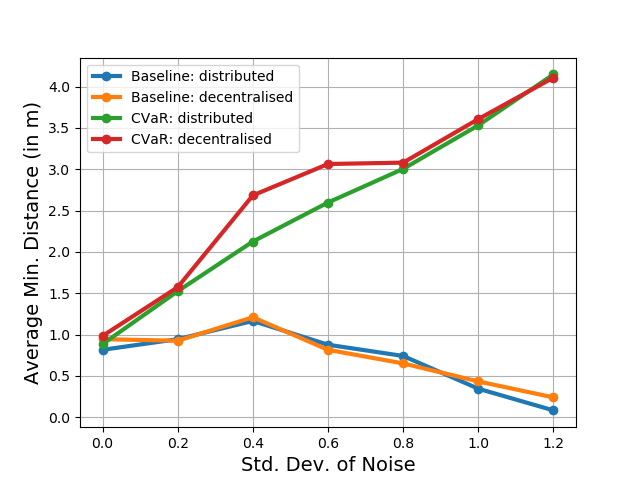}  
\caption{Avgerage minimum distance with increasing noise values (for $\alpha=0.05$ and $0.001$).}
\label{fig:noise_dist}
\end{minipage}
\end{figure*}

\subsection{Gazebo Simulations}

We validate the proposed approach in realistic Gazebo simulations with six agents. Each agent is modeled as a prism of size $1.5 \times 1.5 \times 1$ m$^3$. We fix the value of $\alpha = 0.1$ and $\theta = 0.001$, with $10$ samples without any additional noise. Six agents in a rectangular formation must reach the opposite side of the rectangle. Figure \ref{fig:hexa_3d} shows the trajectories of all the agents for this task. Further videos from Gazebo simulations are included in the supplementary video. 

\subsection{Comparison with the baseline}
In this subsection, we compare our formulation with the baseline deterministic MPC solutions in both distributed and decentralised settings with two agents. In the distributed setting, each agent has the access to the optimal MPC trajectories of other agents and solves a deterministic MPC problem avoiding collision with the predicted trajectories. In the decentralised setting, an agent solves a deterministic MPC problem avoiding collision with the predicted trajectories of others under a constant velocity assumption. These baseline solutions are compared with the data-driven distributionally robust CVaR constrained solutions. To increase the level of uncertainty, we add Gaussian noise to the perceived states and predictions of other agents at each time step. We use the most recent samples based on user-specified sample size, with $\alpha=0.05$ and $\theta=0.001$. Fifty simulations with each parameter configuration are conducted.

As evident from Figure \ref{fig:noise_percent}, collisions occur for the baseline MPC as we increase the noise levels; with the collision percentage being higher for the decentralised case which incorrectly predicts the future positions of other agents based on a constant velocity assumption. The proposed formulation (CVaR-MPC) does not give collision even for very high noise standard deviation in any of our simulations. In Figure \ref{fig:noise_dist}, average of minimum distance between agents with increasing noise levels is plotted. We observe that CVaR-MPC takes a more risk-averse approach, causing higher separation as the level of uncertainty increases. In contrast, baseline approaches lead to a higher proportion of collisions leading to a smaller average minimum distance. Thus, the distributionally robust approach enables us to robustify MPC solutions even with a relatively small samples size. 

\subsection{Computation Time}

Table \ref{tab_time} shows the computation time for different values of sample size for CVaR constraints and prediction horizon of the MPC. This is obtained from the previous numerical simulation setting for the two-agent case. We observe that most of the configurations have computation the time less than 0.1s or 100ms, which is the step size of MPC optimization. The mean and standard deviation of computation time is higher for increasing time horizon and sample size. 

\newcolumntype{C}{>{$}c<{$}}

\begin{table}[!htbp]
\centering
\caption{Computation Time (in milliseconds)}
\label{tab_time}
\begin{tabular}{|p{0.2\linewidth}|C|C|C|}
\hline 
Time Horizon Sample Size & 10 & 20 & 30 
\\\hline\hline
5 & 19\pm2 ms & 40\pm13 ms & 63\pm18 ms 
\\\hline
10 & 27\pm8 ms & 60\pm18 ms & 102\pm76 ms
\\\hline
20 & 41\pm12 ms & 113\pm35 ms & 184\pm127 ms
\\\hline

\end{tabular}
\end{table}

\vspace{-4mm}

\section{Conclusion}
\label{conclusion}
We presented a novel data-driven risk sensitive collision avoidance constraint formulation for safe multi-robot navigation in both distributed and decentralised settings. The proposed approach robustifies MPC solutions against errors in the predictions of surrounding objects in terms of distributional robustness guaranteed by the Wasserstein metric by leveraging data that is collected online during execution. CVaR-based risk constraints capture the proximity to the uncertain polyhedral obstacles and provides the ability for the user to dictate the risk-appetite of the robot. The performance of the proposed approach was examined via numerical simulations with multiple aerial robots, the influence of risk tolerance level and size of the ambiguity sets was illustrated in detail, and further validated on realistic Gazebo simulations. The computation time of the resulting MPC problem is sufficiently small to be deployed in practice. In future, we aim to explore other classes of ambiguity sets for this task that may reduce the computational burden while preserving the desired robustness properties.

\clearpage

\bibliographystyle{IEEEtran}
\bibliography{root}

\begin{thebibliography}{10}
\providecommand{\url}[1]{#1}
\csname url@samestyle\endcsname
\providecommand{\newblock}{\relax}
\providecommand{\bibinfo}[2]{#2}
\providecommand{\BIBentrySTDinterwordspacing}{\spaceskip=0pt\relax}
\providecommand{\BIBentryALTinterwordstretchfactor}{4}
\providecommand{\BIBentryALTinterwordspacing}{\spaceskip=\fontdimen2\font plus
\BIBentryALTinterwordstretchfactor\fontdimen3\font minus
  \fontdimen4\font\relax}
\providecommand{\BIBforeignlanguage}[2]{{%
\expandafter\ifx\csname l@#1\endcsname\relax
\typeout{** WARNING: IEEEtran.bst: No hyphenation pattern has been}%
\typeout{** loaded for the language `#1'. Using the pattern for}%
\typeout{** the default language instead.}%
\else
\language=\csname l@#1\endcsname
\fi
#2}}
\providecommand{\BIBdecl}{\relax}
\BIBdecl

\bibitem{gans2021cooperative}
N.~R. Gans and J.~G. Rogers, ``Cooperative multirobot systems for military
  applications,'' \emph{Current Robotics Reports}, vol.~2, no.~1, pp. 105--111,
  2021.

\bibitem{queralta2020collaborative}
J.~P. Queralta, J.~Taipalmaa, B.~C. Pullinen, V.~K. Sarker, T.~N. Gia,
  H.~Tenhunen, M.~Gabbouj, J.~Raitoharju, and T.~Westerlund, ``Collaborative
  multi-robot search and rescue: Planning, coordination, perception, and active
  vision,'' \emph{IEEE Access}, vol.~8, pp. 191\,617--191\,643, 2020.

\bibitem{malik2021collaborative}
S.~Malik, M.~A. Khan, and H.~El-Sayed, ``Collaborative autonomous driving—a
  survey of solution approaches and future challenges,'' \emph{Sensors},
  vol.~21, no.~11, p. 3783, 2021.

\bibitem{rouvcek2019darpa}
T.~Rou{\v{c}}ek, M.~Pecka, P.~{\v{C}}{\'\i}{\v{z}}ek,
  T.~Pet{\v{r}}{\'\i}{\v{c}}ek, J.~Bayer, V.~{\v{S}}alansk{\`y}, D.~He{\v{r}}t,
  M.~Petrl{\'\i}k, T.~B{\'a}{\v{c}}a, V.~Spurn{\`y} \emph{et~al.}, ``Darpa
  subterranean challenge: Multi-robotic exploration of underground
  environments,'' in \emph{International Conference on Modelling and Simulation
  for Autonomous Systems}.\hskip 1em plus 0.5em minus 0.4em\relax Springer,
  2019, pp. 274--290.

\bibitem{jiang2019multi}
Y.~Jiang, H.~Yedidsion, S.~Zhang, G.~Sharon, and P.~Stone, ``Multi-robot
  planning with conflicts and synergies,'' \emph{Autonomous Robots}, vol.~43,
  no.~8, pp. 2011--2032, 2019.

\bibitem{pinkam2016robot}
N.~Pinkam, F.~Bonnet, and N.~Y. Chong, ``Robot collaboration in warehouse,'' in
  \emph{2016 16th International Conference on Control, Automation and Systems
  (ICCAS)}.\hskip 1em plus 0.5em minus 0.4em\relax IEEE, 2016, pp. 269--272.

\bibitem{o2021there}
J.~O'Malley, ``There's no business like drone business [drone light shows],''
  \emph{Engineering \& Technology}, vol.~16, no.~4, pp. 72--79, 2021.

\bibitem{huang2019collision}
S.~Huang, R.~S.~H. Teo, and K.~K. Tan, ``Collision avoidance of multi unmanned
  aerial vehicles: A review,'' \emph{Annual Reviews in Control}, vol.~48, pp.
  147--164, 2019.

\bibitem{luis2020online}
C.~E. Luis, M.~Vukosavljev, and A.~P. Schoellig, ``Online trajectory generation
  with distributed model predictive control for multi-robot motion planning,''
  \emph{IEEE Robotics and Automation Letters}, vol.~5, no.~2, pp. 604--611,
  2020.

\bibitem{kamel2017robust}
M.~Kamel, J.~Alonso-Mora, R.~Siegwart, and J.~Nieto, ``Robust collision
  avoidance for multiple micro aerial vehicles using nonlinear model predictive
  control,'' in \emph{2017 IEEE/RSJ International Conference on Intelligent
  Robots and Systems (IROS)}.\hskip 1em plus 0.5em minus 0.4em\relax IEEE,
  2017, pp. 236--243.

\bibitem{zhang2020optimization}
X.~Zhang, A.~Liniger, and F.~Borrelli, ``Optimization-based collision
  avoidance,'' \emph{IEEE Transactions on Control Systems Technology}, vol.~29,
  no.~3, pp. 972--983, 2020.

\bibitem{batkovic2020robust}
I.~Batkovic, U.~Rosolia, M.~Zanon, and P.~Falcone, ``A robust scenario mpc
  approach for uncertain multi-modal obstacles,'' \emph{IEEE Control Systems
  Letters}, vol.~5, no.~3, pp. 947--952, 2020.

\bibitem{soloperto2019collision}
R.~Soloperto, J.~K{\"o}hler, F.~Allg{\"o}wer, and M.~A. M{\"u}ller, ``Collision
  avoidance for uncertain nonlinear systems with moving obstacles using robust
  model predictive control,'' in \emph{2019 18th European Control Conference
  (ECC)}.\hskip 1em plus 0.5em minus 0.4em\relax IEEE, 2019, pp. 811--817.

\bibitem{katriniok2018distributed}
A.~Katriniok, S.~Kojchev, E.~Lefeber, and H.~Nijmeijer, ``Distributed scenario
  model predictive control for driver aided intersection crossing,'' in
  \emph{2018 European Control Conference (ECC)}.\hskip 1em plus 0.5em minus
  0.4em\relax IEEE, 2018, pp. 1746--1752.

\bibitem{dixit2022risk}
A.~Dixit, M.~Ahmadi, and J.~W. Burdick, ``Risk-averse receding horizon motion
  planning,'' \emph{arXiv preprint arXiv:2204.09596}, 2022.

\bibitem{cheng2017decentralized}
H.~Cheng, Q.~Zhu, Z.~Liu, T.~Xu, and L.~Lin, ``Decentralized navigation of
  multiple agents based on orca and model predictive control,'' in \emph{2017
  IEEE/RSJ International Conference on Intelligent Robots and Systems
  (IROS)}.\hskip 1em plus 0.5em minus 0.4em\relax IEEE, 2017, pp. 3446--3451.

\bibitem{hakobyan2019risk}
A.~Hakobyan, G.~C. Kim, and I.~Yang, ``Risk-aware motion planning and control
  using cvar-constrained optimization,'' \emph{IEEE Robotics and Automation
  letters}, vol.~4, no.~4, pp. 3924--3931, 2019.

\bibitem{castillo2020real}
M.~Castillo-Lopez, P.~Ludivig, S.~A. Sajadi-Alamdari, J.~L. Sanchez-Lopez,
  M.~A. Olivares-Mendez, and H.~Voos, ``A real-time approach for
  chance-constrained motion planning with dynamic obstacles,'' \emph{IEEE
  Robotics and Automation Letters}, vol.~5, no.~2, pp. 3620--3625, 2020.

\bibitem{zhu2019chance}
H.~Zhu and J.~Alonso-Mora, ``Chance-constrained collision avoidance for mavs in
  dynamic environments,'' \emph{IEEE Robotics and Automation Letters}, vol.~4,
  no.~2, pp. 776--783, 2019.

\bibitem{de2021scenario}
O.~de~Groot, B.~Brito, L.~Ferranti, D.~Gavrila, and J.~Alonso-Mora,
  ``Scenario-based trajectory optimization in uncertain dynamic environments,''
  \emph{IEEE Robotics and Automation Letters}, vol.~6, no.~3, pp. 5389--5396,
  2021.

\bibitem{gao2021risk}
Y.~Gao, F.~J. Jiang, L.~Xie, and K.~H. Johansson, ``Risk-aware optimal control
  for automated overtaking with safety guarantees,'' \emph{IEEE Transactions on
  Control Systems Technology}, 2021.

\bibitem{firoozi2020distributed}
R.~Firoozi, L.~Ferranti, X.~Zhang, S.~Nejadnik, and F.~Borrelli, ``A
  distributed multi-robot coordination algorithm for navigation in tight
  environments,'' \emph{arXiv preprint arXiv:2006.11492}, 2020.

\bibitem{dai2022distributed}
L.~Dai, Y.~Hao, H.~Xie, Z.~Sun, and Y.~Xia, ``Distributed robust mpc for
  nonholonomic robots with obstacle and collision avoidance,'' \emph{Control
  Theory and Technology}, vol.~20, no.~1, pp. 32--45, 2022.

\bibitem{arul2020dcad}
S.~H. Arul and D.~Manocha, ``Dcad: Decentralized collision avoidance with
  dynamics constraints for agile quadrotor swarms,'' \emph{IEEE Robotics and
  Automation Letters}, vol.~5, no.~2, pp. 1191--1198, 2020.

\bibitem{berg2011reciprocal}
J.~v.~d. Berg, S.~J. Guy, M.~Lin, and D.~Manocha, ``Reciprocal n-body collision
  avoidance,'' in \emph{Robotics research}.\hskip 1em plus 0.5em minus
  0.4em\relax Springer, 2011, pp. 3--19.

\bibitem{park2020online}
J.~Park and H.~J. Kim, ``Online trajectory planning for multiple quadrotors in
  dynamic environments using relative safe flight corridor,'' \emph{IEEE
  Robotics and Automation Letters}, vol.~6, no.~2, pp. 659--666, 2020.

\bibitem{gopalakrishnan2017prvo}
B.~Gopalakrishnan, A.~K. Singh, M.~Kaushik, K.~M. Krishna, and D.~Manocha,
  ``Prvo: Probabilistic reciprocal velocity obstacle for multi robot navigation
  under uncertainty,'' in \emph{2017 IEEE/RSJ International Conference on
  Intelligent Robots and Systems (IROS)}.\hskip 1em plus 0.5em minus
  0.4em\relax IEEE, 2017, pp. 1089--1096.

\bibitem{arul2021swarmcco}
S.~H. Arul and D.~Manocha, ``Swarmcco: Probabilistic reactive collision
  avoidance for quadrotor swarms under uncertainty,'' \emph{IEEE Robotics and
  Automation Letters}, vol.~6, no.~2, pp. 2437--2444, 2021.

\bibitem{zhang2021receding}
X.~Zhang, J.~Ma, Z.~Cheng, S.~Huang, and T.~H. Lee, ``Receding horizon motion
  planning for multi-agent systems: A velocity obstacle based probabilistic
  method,'' \emph{arXiv preprint arXiv:2103.12968}, 2021.

\bibitem{wang2022risk}
Y.~Wang and M.~P. Chapman, ``Risk-averse autonomous systems: A brief history
  and recent developments from the perspective of optimal control,''
  \emph{Artificial Intelligence}, p. 103743, 2022.

\bibitem{hakobyan2021wasserstein}
A.~Hakobyan and I.~Yang, ``Wasserstein distributionally robust motion control
  for collision avoidance using conditional value-at-risk,'' \emph{IEEE
  Transactions on Robotics}, vol.~38, no.~2, pp. 939--957, 2021.

\bibitem{hota2019data}
A.~R. Hota, A.~Cherukuri, and J.~Lygeros, ``Data-driven chance constrained
  optimization under wasserstein ambiguity sets,'' in \emph{2019 American
  Control Conference (ACC)}.\hskip 1em plus 0.5em minus 0.4em\relax IEEE, 2019,
  pp. 1501--1506.

\bibitem{sabatino2015quadrotor}
F.~Sabatino, ``Quadrotor control: modeling, nonlinearcontrol design, and
  simulation,'' 2015.

\bibitem{wachter2006implementation}
A.~W{\"a}chter and L.~T. Biegler, ``On the implementation of an interior-point
  filter line-search algorithm for large-scale nonlinear programming,''
  \emph{Mathematical programming}, vol. 106, no.~1, pp. 25--57, 2006.

\bibitem{lucia2017rapid}
S.~Lucia, A.~T{\u{a}}tulea-Codrean, C.~Schoppmeyer, and S.~Engell, ``Rapid
  development of modular and sustainable nonlinear model predictive control
  solutions,'' \emph{Control Engineering Practice}, vol.~60, pp. 51--62, 2017.

\bibitem{Andersson2019}
J.~A.~E. Andersson, J.~Gillis, G.~Horn, J.~B. Rawlings, and M.~Diehl,
  ``{CasADi} -- {A} software framework for nonlinear optimization and optimal
  control,'' \emph{Mathematical Programming Computation}, vol.~11, no.~1, pp.
  1--36, 2019.

\bibitem{furrer2016rotors}
F.~Furrer, M.~Burri, M.~Achtelik, and R.~Siegwart, ``Rotors—a modular gazebo
  mav simulator framework,'' in \emph{Robot operating system (ROS)}.\hskip 1em
  plus 0.5em minus 0.4em\relax Springer, 2016, pp. 595--625.

\end{thebibliography}

\end{document}